\documentclass{ecai2014}
\usepackage{times}
\usepackage{graphicx}
\usepackage{latexsym}
\usepackage{algorithm}
\usepackage{algorithmic}

\begin{document}

\title{Consistency-based Merging of Variability Models}

\author{Mathias Uta\institute{Siemens Erlangen,
Germany, email: mathias.uta@siemens.com} \and {Alexander Felfernig}\institute{Graz University of Technology,
Austria, email: alexander.felfernig@ist.tugraz.at, spoecklberger.johannes@student.tugraz.at}\\ \and Gottfried Schenner\institute{Siemens AG,
Austria, email: gottfried.schenner@siemens.com} \and Johannes Sp\"ocklberger$^2$}

\maketitle

\begin{abstract}
  Globally operating enterprises selling large and complex products and services often have to deal with situations where variability models are locally developed to take into account the requirements of local markets. For example, cars sold on the U.S. market are represented by variability models in some or many aspects different from European ones. In order to support global variability management processes, variability models and the underlying knowledge bases often need to be integrated. This is a challenging task since an integrated knowledge base should not produce results which are different from those produced by the individual knowledge bases. In this paper, we introduce an approach to variability model integration that is based on the concepts of contextual modeling and conflict detection. We present the underlying concepts and the results of a corresponding performance analysis.
\end{abstract}


\maketitle              
\section{Introduction}\label{introduction}
Configuration \cite{felfernighotzbagleytiihonen2014,Stumptner1997} is one of the most successful applications of Artificial Intelligence technologies applied in domains such as telecommunication switches, financial services, furniture, and software components. In many cases, configuration knowledge bases are represented in terms of variability models such as feature models that provide an intuitive way of representing variability properties of complex systems \cite{Kang1990,Czarnecki2005}. Starting with rule-based approaches, formalizations of variability models have been transformed into model-based knowledge representations which are more applicable for the handling of large and complex knowledge bases, for example, in terms of knowledge base maintainability and expressivity of complex constraints \cite{Benavides2010,felfernighotzbagleytiihonen2014}. Examples of model-based knowledge representations are constraint-based representations \cite{Tsang1993}, description logic and answer set programming (ASP) \cite{felfernighotzbagleytiihonen2014}. Besides variability reasoning for single users, latest research also shows how to deal with scenarios where groups of users are completing a configuration task \cite{Felfernig2018}. In this paper, we focus on single user scenarios where variability models are represented as a constraint satisfaction problem (CSP) \cite{Benavides2005,Tsang1993}.

There exist a couple of approaches dealing with the issue of integrating knowledge bases. First, \emph{knowledge base alignment} is the process of identifying relationships between concepts in different knowledge bases, for example, classes describe the same concept but have different class names (and/or attribute names). Approaches supporting the alignment of knowledge bases are relevant in scenarios where numerous and large knowledge bases have to be integrated (see, for example, \cite{Galarraga2013}). Ardissono et al. \cite{Ardissono2003} introduce an approach to \emph{distributed configuration} where individual knowledge bases are integrated into a distributed configuration process in which individual configurators are responsible for configuring individual parts of a complex product or service. The underlying assumption is that individual knowledge bases are consistent and that there are no (or only a low number of) dependencies between the given knowledge bases. 

The \emph{merging of knowledge bases} is related to the task of exploiting various merging operators to different belief sets \cite{Delgrande2007,Liberatore1998}. For example, Delgrande and Schaub \cite{Delgrande2007} introduce a consistency-based merging approach where the result of a merging process is a maximum consistent set of logical formulas representing the union of the individual knowledge bases. In the line of existing consistency-based analysis approaches, the resulting knowledge bases represent a logical union of the original knowledge bases that omits minimal sets of logical sentences inducing an inconsistency \cite{84_Reiter1987}.  \emph{Contextual modeling} \cite{felfernig2000} is related to the task of decentralizing variability knowledge related  development and maintenance tasks. 

Approaches to merging \emph{feature models} represented on a graphical level on the basis of merging rules have been introduced, for example, in \cite{Broek2010,Segura2007}. In this context, feature models including specific constraint types such as \emph{requires} and \emph{excludes}, are merged in a semantics-preserving fashion. Compared to our approach, the merging of variability models introduced in \cite{Broek2010,Segura2007} is restricted to specific constraint types and does not take into account redundancy. 

Our approach provides a generalization to existing approaches especially due to the generalization to arbitrary constraint types and redundancy-free knowledge bases as a result of the merge operation. We propose an approach to the \emph{merging of variability models} (represented as constraint satisfaction problems) which guarantees semantics preservation, i.e., the union of the solutions determined by individual constraint solvers (configurators) is equivalent to the solution space of the integrated variability model (knowledge base). In this context, we assume that the knowledge bases to be integrated  (1) are consistent and (2) use the same variable names for representing individual item properties (knowledge base alignment issues are beyond the scope of this paper).

The contributions of this paper are the following. (1) We provide a short analysis of existing approaches to knowledge base integration and point out specific properties of variability model integration scenarios that require alternative approaches. (2) We introduce a new approach to variability knowledge integration which is based on the concepts of contextualization and conflict detection. (3) We show the applicability of a our approach on the basis of a performance analysis. 

The remainder of this paper is organized as follows. First, we introduce a working example from the automotive domain (see Section \ref{workingexample}). On the basis of this example, we introduce our approach to variability model integration (merging) in Section \ref{integratingconfigurationknowledgebases}. In Section \ref{performanceevaluation}, we present a performance evaluation. Section \ref{threatstovalidity} includes a discussion of threats to validity of the presented merging approach. The paper is concluded in Section \ref{conclusionsandfuturework} with a discussion of issues for future work.

\section{Example Variability Models}\label{workingexample}
In the following, we introduce a working example which will serve as a basis for the discussion of our approach to knowledge integration (Section \ref{integratingconfigurationknowledgebases}). Let us assume the existence of two different variability models. For the purpose of our example, we introduce two car configuration knowledge bases represented as a constraint satisfaction problem. One car configuration knowledge base is assumed to be defined for the U.S. market and one for the German market. For simplicity, we assume that (1) both knowledge bases are represented as a constraint satisfaction problem (CSP) \cite{Tsang1993} and (2) that both knowledge bases operate on the same set of variables and corresponding domain definitions.\footnote{We are aware of the fact that this assumption does not hold for real-world scenarios in general. However, we consider tasks of concept matching as an upstream task we do not take into account when integrating knowledge bases on a formal level.} Our two knowledge bases consisting of variable definitions and corresponding constraints \{$CKB_{us}$, $CKB_{ger}$\} are the following. 

\begin{itemize}
\item{$CKB_{us}$: \{country(US), type(combi, limo, city, suv), color(white, black), engine(1l, 1.5l, 2l), couplingdev(yes,no), fuel(electro, diesel, gas, hybrid), service(15k, 20k, 25k), $c_{1us}:fuel \neq hybrid$, $c_{2us}:fuel = electro \rightarrow coupling-$ $dev = no$, $c_{3us}:fuel = diesel \rightarrow color = black$\}}
\item{$CKB_{ger}$: \{country(GER), type(combi, limo, city, suv), color(white, black), engine(1l, 1.5l, 2l), couplingdev(yes,no), fuel(electro, diesel, gas, hybrid), service(15k, 20k, 25k), $c_{1ger}:fuel \neq gas$, $c_{2ger}:fuel = electro \rightarrow couplingdev = no$, $c_{3ger}:fuel = diesel \rightarrow type \neq city$\}}
\end{itemize}

In these knowledge bases, we denote the variable \emph{country} as contextual variable since it is used to specify the country a configuration belongs to but is not directly associated with a specific component of the car. Table \ref{solutionspacesindividual} shows a summary of the solution spaces (in terms of the number of potential solutions) that are associated with the country-specific knowledge bases $CKB_{us}$ and $CKB_{ger}$. For simplicity, we kept the number of constraints the same in both knowledge bases, however, the integration concepts introduced in Section \ref{integratingconfigurationknowledgebases} are also applicable to knowledge bases with differing numbers of constraints.

\vspace{-0.15cm}

\begin{table}[!ht]
\caption{Solution spaces of individual knowledge bases.} \label{solutionspacesindividual}
\centering{}\begin{tabular}{|c|c|c|c|c|c|} 
\hline 
Knowledge base      	& \#constraints 	& \#solutions  	        \tabularnewline
\hline
\hline
$CKB_{us}$ 				& 3 		        & 288 		       \tabularnewline
\hline
$CKB_{ger}$ 			& 3 		        & 324 		       \tabularnewline
\hline
\end{tabular}  
\end{table}

\section{Merging Variability Models}\label{integratingconfigurationknowledgebases}
In this section, we introduce our approach to merge variability models represented as constraint satisfaction problems (CSPs) \cite{Tsang1993}. Our approach is based on the assumption that the constraints of the two original knowledge bases $CKB_1$ and $CKB_2$ are contextualized, i.e., each constraint of knowledge base $CKB_1$ gets contextualized on the basis of predefined contextualization variables. For example: assuming a context variable \emph{country}(US,GER), each constraint $c_{[i]us}$ of the US knowledge base is contextualized with (transformed into) $country = US \rightarrow (c_{[i]us})$. Constraint $c_{1us}: fuel \neq hybrid$ would be translated into $c_{1us'}: country = US \rightarrow (fuel \neq hybrid)$. $CKB_{us}$ and $CKB_{ger}$ have been transformed into their contextualized variants  $CKB_{us}'$ and $CKB_{ger}'$ where $CKB_{us}' \cup CKB_{ger}' = CKB'$.

\begin{itemize}
\item{$CKB_{us}'$: \{country(US), type(combi, limo, city, suv), color(white, black), engine(1l, 1.5l, 2l), couplingdev(yes,no), fuel(electro, diesel, gas, hybrid), service(15k, 20k, 25k), $c_{1us}': country = US \rightarrow (fuel \neq hybrid$), $c_{2us}':country = US \rightarrow (fuel = electro \rightarrow couplingdev = no$), $c_{3us}:country = US \rightarrow (fuel = diesel \rightarrow color = black$)\}}
\item{$CKB_{ger}'$: \{country(GER), type(combi, limo, city, suv), color(white, black), engine(1l, 1.5l, 2l), couplingdev(yes,no), fuel(electro, diesel, gas, hybrid), service(15k, 20k, 25k), $c_{1ger}': country = GER \rightarrow (fuel \neq gas$), $c_{2ger}': country = GER \rightarrow (fuel = electro \rightarrow couplingdev = no$), $c_{3ger}': country = GER \rightarrow (fuel = diesel \rightarrow type \neq city$)\}}
\end{itemize}

The solution spaces of the contextualized knowledge bases $CKB_{us}'$ and $CKB_{ger}'$ are shown in Table \ref{solutionspacescontextualized}. They have the same solution spaces as $CKB_{us}$ and $CKB_{ger}$.

\begin{table}[!ht]
\caption{Solution spaces when merging knowledge bases.} \label{solutionspacescontextualized}
\centering{}\begin{tabular}{|c|c|c|c|c|c|} 
\hline 
Knowledge base      	 	    & \#solutions  	        \tabularnewline
\hline
\hline
$CKB_{us}'$ 				    & 288 		       \tabularnewline
\hline
$CKB_{ger}'$ 			        & 324 		       \tabularnewline
\hline
$CKB' = CKB_{us}' \cup CKB_{ger}'$ 			                & 612 		       \tabularnewline
\hline
$CKB_{us}'\cap CKB_{ger}'$   	& 126 		       \tabularnewline
\hline
\end{tabular}  
\end{table}

\vspace{0.25cm}

On the basis of such a contextualization, we are able to preserve the consistency and semantics of the two original knowledge bases in the sense that (1) the solution space ($CKB_1$) is equivalent to the solution space ($CKB_1'$), (2) the solution space ($CKB_2$) is equivalent to the solution space ($CKB_2'$), and (3) the solution space ($CKB_1$) $\cup$ solution space ($CKB_2$) is equivalent to the solution space ($CKB_1' \cup CKB_2' = CKB'$).

Based on this representation, we are able to (1) get rid of contextualizations (see line $7$ of Algorithm 1) that are not needed in the integrated version of the two original configuration knowledge bases and (2)  delete redundant constraints (see line 15 of Algorithm 1). In Line $7$ it is checked whether a contextualization is needed for the constraint $c$ ($c$ is the decontextualized version of $c'$). If the negation of $c$ is consistent with the union of the contextualized knowledge bases, solutions exist that support $\neg c$. Consequently, $c$ must remain contextualized. Otherwise, the contextualization is not needed and $c$ is added to the resulting knowledge base -- with this, it replaces $c'$, i.e., the corresponding contextualized constraint. 

Each constraint in the resulting knowledge base $CKB$ (the decontextualized knowledge base) is thereafter checked with regard to redundancy (see Line $15$). A constraint $c$ is regarded as redundant if $CKB - \{c\}$ is inconsistent with $\neg c$. In this case, $c$ does not reduce the search space and thus can be deleted from $CKB$ -- it is redundant with regard to $CKB$.

\vspace{0.25cm}

\algsetup{ linenosize={\small } }
\begin{algorithm}[ht]
 \caption{\textsc{CKB-Merge}($CKB_1', CKB_2'$)$:CKB$}
\label{alg:Sequential} 
\begin{algorithmic}[1]
\STATE \COMMENT{$CKB_{1',2'}$:  two contextualized and consistent configuration knowledge bases}
\STATE \COMMENT{$c'$:  a contextualized version of constraint c}
\STATE \COMMENT{$CKB$: knowledge base resulting from merge operation} 
\STATE $CKB ~ \gets$ $\emptyset$;
\STATE $CKB' ~ \gets$ $CKB_{1'} \cup CKB_{2'}$;
\FORALL {$c'$ $\in$ $CKB'$}
\IF {$inconsistent(\{\neg c\} \cup CKB' \cup CKB$)}
\STATE $CKB ~ \gets$ $CKB ~ \cup ~ \{c\};$ 
\ELSE 
\STATE $CKB ~ \gets$ $CKB ~ \cup ~ \{c'\};$
\ENDIF
\STATE $CKB' ~ \gets$ $CKB' ~ - ~ \{c'\};$
\ENDFOR
\FORALL {$c$ $\in$ $CKB$}
\IF {$inconsistent((CKB - \{c\}) \cup \{\neg c\})$} \STATE $CKB ~ \gets CKB ~ - ~ \{c\};$
\ENDIF
\ENDFOR
\STATE $return ~ CKB;$
\end{algorithmic} 
\end{algorithm}

\vspace{0.25cm}

The knowledge base $CKB$ resulting from applying Algorithm 1 to the individual knowledge bases $CKB_{us}'$ and $CKB_{ger}'$ looks like as follows. In $CKB$, constraint $c_{2us}'$ is represented in a decontextualized fashion since the context information is not needed. Furthermore, constraint $c_{2ger}'$ has been deleted since it is redundant.

\begin{itemize}
\item{$CKB$: \{country(US, GER), type(combi, limo, city, suv), color(white, black), engine(1l, 1.5l, 2l), couplingdev(yes,no), fuel(electro, diesel, gas, hybrid), service(15k, 20k, 25k), $c_{1us}': country = US \rightarrow (fuel \neq hybrid$), $c_{2us}: fuel = electro \rightarrow couplingdev = no$, $c_{3us}':country = US \rightarrow (fuel = diesel \rightarrow color = black$), $c_{1ger}': country = GER \rightarrow (fuel \neq gas$),  $c_{3ger}': country = GER \rightarrow (fuel = diesel \rightarrow type \neq city$)\}}
\end{itemize}

\section{Performance Evaluation}\label{performanceevaluation}

In this section, we  discuss the results of an initial analysis we have conducted to evaluate \textsc{CKB-Merge} (Algorithm 1). For this analysis, we applied 10 different synthesized variability models $CKB'$ ($CKB' = CKB_1' \cup CKB_2'$) represented as constraint satisfaction problems \cite{Tsang1993}) that differ individually in terms of the number of constraints (\#constraints) and the degree of contextualization (expressed as percentages in Tables \ref{runtimeckbmerge} and \ref{runtimeckbs}). In order to take into account deviations in time measurements, we repeated each experimental setting 10 times where in each repetition cycle the constraints in the individual (contextualized) knowledge bases $CKB'$ were ordered randomly.

The number of consistency checks needed for decontextualization is linear in terms of the number of constraints in $CKB'$. A performance evaluation of \textsc{CKB-Merge} with different knowledge base sizes and degrees of contextualized constraints in $CKB$ is depicted in Table \ref{runtimeckbmerge}. In \textsc{CKB-Merge}, the runtime (measured in terms of milliseconds needed by the constraint solver\footnote{For the purposes of our evaluation we generated variability models represented as constraint satisfaction problems formulated using the \textsc{Choco} constraint solver -- www.choco-solver.org.} to find a solution) increases with the number of constraints in $CKB'$ and decreases with the number of contextualized constraints in $CKB$. The increase in efficiency can be explained by the fact that a higher degree of contextualization includes more situations where the inconsistency check in Line 7 (Algorithm 1) terminates earlier (a solution has been found) compared to situations where no solution could be found. In addition, Table \ref{runtimeckbs} indicates that the performance of solution search does not differ depending on the degree of contextualization in the resulting knowledge base $CKB$.

Consequently, integrating individual variability models can trigger the following improvements. (1) De-contextualization in $CKB$ can lead to less cognitive efforts when adapting / extending knowledge bases (due to a potentially lower number of constraints and a lower degree of contextualization). (2) Reducing the overall number of constraints in $CKB$ can also improve runtime performance of the resulting integrated knowledge base.

\begin{table}
\caption{Avg. runtime (\emph{msec}) of \textsc{CKB-Merge} measured with different \emph{knowledge base sizes} ($CKB'$) and shares of contextualized constraints in  $CKB$ (10-50\% contextualization).} \label{runtimeckbmerge}
\centering{}\begin{tabular}{|c|c|c|c|c|c|c|c|c|c|} 
\hline 
$CKB'$  & \#constraints      	& 10\% 	& 20\%   & 30\%	 & 40\%  & 50\%     \tabularnewline
\hline
\hline
1             & 10 				    &  	749 & 219 & 195 & 118 & 97          \tabularnewline
\hline
2             & 20 				    &  	559 & 653 & 666 & 679 & 487                     \tabularnewline
\hline
3             & 30 				    &  	1541 & 813 & 644 & 588 & 664                    \tabularnewline
\hline
4             & 40 				    &  	1888 & 1541 & 1345 & 1177 & 1182                   \tabularnewline
\hline
5             & 50 				    &  	3773 & 3324 & 3027 & 3171 & 2643                  \tabularnewline
\hline
6             & 60 				    &  	5376 & 4458 & 4425 & 3304 & 3056                    \tabularnewline
\hline
7             & 70 				    &  	7300 & 6912 & 7362 & 5619 & 4896                    \tabularnewline
\hline
8             & 80 				    &  	10795 & 8793 & 7580 & 6821 & 5909                    \tabularnewline
\hline
9             & 90 				    &  	13365 & 11770 & 10103 & 8916 & 7831                   \tabularnewline
\hline
10             & 100 				&  	15992 & 14443 & 14679 & 12417 & 11066                  \tabularnewline
\hline
\end{tabular}  
\end{table}

\begin{table}
\caption{Avg. runtime (\emph{msec}) of the merged configuration knowledge bases (CKB) measured with different \emph{knowledge base sizes} ($CKB'$) and shares of contextualized constraints in  $CKB$ (10-50\% contextualization).} \label{runtimeckbs}
\centering{}\begin{tabular}{|c|c|c|c|c|c|c|c|c|c|} 
\hline 
$CKB'$  &   \#constraints &  10\% 	& 20\%   & 30\%	 & 40\%  & 50\%     \tabularnewline
\hline
\hline
1             & 10 				    &  		    244 & 159 & 203 & 167 & 274            \tabularnewline
\hline
2             & 20				    &  		    305 & 230 & 250 & 362 & 271            \tabularnewline
\hline
3             & 30 				    &  		    310 & 378 & 251 & 426 & 243            \tabularnewline
\hline
4             & 40 				    &  		    425 & 453 & 522 & 502 & 563            \tabularnewline
\hline
5             & 50 				    &  		    500 & 640 & 603 & 637 & 657            \tabularnewline
\hline
6            & 60  				    &  		    881 & 728 & 899 & 801 & 698            \tabularnewline
\hline
7            & 70  				    &  		    830 & 778 & 802 & 888 & 876            \tabularnewline
\hline
8            & 80  				    &  		    917 & 1054 & 1011 & 848 & 1030            \tabularnewline
\hline
9           & 90   				    &  		    1017 & 1117 & 1042 & 960 & 667            \tabularnewline
\hline
10          & 100   				&  		    1387 & 1363 & 1297 & 1297 & 1308            \tabularnewline
\hline
\end{tabular}  
\end{table}

\section{Threats to Validity}\label{threatstovalidity}
The main threat to (external) validity is the overall representativeness of the knowledge bases used for evaluating the performance of \textsc{CKB-Merge}. The current evaluation is based on a set of synthesized knowledge bases which do not directly reflect real-world variability models. We want to point out that the major focus of our work is to provide an algorithmic solution that allows semantics-preserving knowledge integration which is a new approach and regarded as the major contribution of our work. The application of \textsc{CKB-Merge} to real-world variability models, i.e., not synthesized ones, is in the focus of our future work.

\section{Conclusions and Future Work}\label{conclusionsandfuturework}

In this paper, we have introduced an approach to the consistency-based merging of variability models represented as constraint satisfaction problems. The approach helps to build semantics-preserving knowledge bases in the sense that the solution space of the resulting knowledge base (result of the merging process) corresponds to the union of the solution spaces of the original knowledge bases. Besides the preservation of the original semantics, our approach also helps to make the resulting knowledge base compact in the sense of deleting redundant constraints and not needed contextual information. The performance of our approach is shown on the basis of a first performance analysis with synthesized configuration knowledge bases. Future work will include the evaluation of our concepts with more complex knowledge bases and the development of alternative merge algorithms with the goal to further improve runtime performance.


%
%
\bibliographystyle{ecai2014}
\bibliography{sigproc}

\end{document}